\documentclass[twoside,11pt]{article}

\usepackage[preprint]{jmlr2e}

\jmlrheading{1}{2022}{1-48}{4/00}{10/00}{zhu00a}{Zhaocheng Zhu, Chence Shi, Zuobai Zhang, Shengchao Liu, Minghao Xu, Xinyu Yuan, Yangtian Zhang, Junkun Chen, Huiyu Cai, Jiarui Lu, Chang Ma, Runcheng Liu, Louis-Pascal Xhonneux, Meng Qu and Jian Tang}

\ShortHeadings{TorchDrug: A Powerful and Flexible Machine Learning Platform for Drug Discovery}{Zhu et al.}
\firstpageno{1}

\usepackage{xspace}
\usepackage{xcolor}
\usepackage[frozencache,cachedir=minted]{minted}
\usepackage{hyperref}
\usepackage{booktabs}
\usepackage{multirow}
\usepackage{adjustbox}
\usepackage[htt]{hyphenat}

\newcommand{\code}[1]{\texttt{#1}}
\newcommand{\library}{\textsc{TorchDrug}\xspace}

\definecolor{mygray}{RGB}{246, 246, 246}

\begin{document}

\title{TorchDrug: A Powerful and Flexible Machine Learning Platform for Drug Discovery}

\author{\name Zhaocheng Zhu\textsuperscript{1,2} \email zhuzhaoc@mila.quebec \\
       \name Chence Shi\textsuperscript{1,2} \email chence.shi@mila.quebec \\
       \name Zuobai Zhang\textsuperscript{1,2} \email zuobai.zhang@mila.quebec \\
       \name Shengchao Liu\textsuperscript{1,2} \email liusheng@mila.quebec \\
       \name Minghao Xu\textsuperscript{1,2} \email minghao.xu@mila.quebec \\
       \name Xinyu Yuan\textsuperscript{3} \email katarinayuan@pku.edu.cn \\
       \name Yangtian Zhang\textsuperscript{4} \email zjhz-zyt@sjtu.edu.cn \\
       \name Junkun Chen\textsuperscript{5} \email chenjk17@mails.tsinghua.edu.cn \\
       \name Huiyu Cai\textsuperscript{1,2} \email huiyu.cai@mila.quebec \\
       \name Jiarui Lu\textsuperscript{1,2} \email jiarui.lu@mila.quebec\\
       \name Chang Ma\textsuperscript{3} \email changma@pku.edu.cn \\
       \name Runcheng Liu\textsuperscript{5} \email lrc18@mails.tsinghua.edu.cn\\
       \name Louis-Pascal Xhonneux\textsuperscript{1,2} \email xhonneul@mila.quebec \\
       \name Meng Qu\textsuperscript{1,2} \email qumeng@mila.quebec \\
       \name Jian Tang\textsuperscript{1,6,7} \email jian.tang@hec.ca \\
       \addr
       \textsuperscript{1}Mila - Quebec AI Institute \\
       \textsuperscript{2}University of Montreal \\
       \textsuperscript{3}Peking University \\
       \textsuperscript{4}Shanghai Jiao Tong University \\
       \textsuperscript{5}Tsinghua University \\
       \textsuperscript{6}HEC Montreal \\
       \textsuperscript{7}CIFAR AI Research Chair
       }
\editor{}

\maketitle

\begin{abstract}

Machine learning has huge potential to revolutionize the field of drug discovery and is attracting increasing attention in recent years. However, lacking domain knowledge (e.g., which tasks to work on), standard benchmarks and data preprocessing pipelines are the main obstacles for machine learning researchers to work in this domain. To facilitate the progress of machine learning for drug discovery, we develop \library, a powerful and flexible machine learning platform for drug discovery built on top of PyTorch. \library benchmarks a variety of important tasks in drug discovery, including molecular property prediction, pretrained molecular representations, de novo molecular design and optimization, retrosynthsis prediction, and biomedical knowledge graph reasoning. State-of-the-art techniques based on geometric deep learning (or graph machine learning), deep generative models, reinforcement learning and knowledge graph reasoning are implemented for these tasks.  \library features a hierarchical interface that facilitates customization from both novices and experts in this domain. Tutorials, benchmark results and documentation are available at \url{https://torchdrug.ai}. Code is released under Apache License 2.0.
\end{abstract}


\section{Introduction}

Drug discovery is a long and costly process, taking on average 10 years and costing 2.5 billion US dollars to develop a new drug~\citep{dimasi2016innovation}. Machine learning has huge potential to accelerate the process of drug discovery by extracting evidence through mining and analyzing data in the biomedical domain (e.g., scientific literature, bioassays, and clinical trials). Recently, machine learning methods have made significant progress in many drug discovery tasks, such as protein structure prediction~\citep{baek2021accurate, jumper2021highly}, molecular property prediction~\citep{duvenaud2015convolutional, hu2019strategies}, de novo molecular design and optimization~\citep{you2018graph, shi2020graphaf} 
, reaction prediction~\citep{jin2017predicting, bradshaw2018generative}, retrosynthesis prediction~\citep{dai2019retrosynthesis, shi2020graph}, and drug repurposing~\citep{wang2020covid, zhao2020biomedical}. 
However, it remains a challenge for machine learning researchers to work in this domain for a few reasons: (1) lacking domain knowledge of what are important tasks in the domain; (2) no standard benchmarks of different methods due to their completely different implementations; (3) the large cost of implementing complicated data preprocessing pipelines for each task.

\begin{figure}[!h]
    \centering
    \vspace{-1em}
    \includegraphics[width=0.68\textwidth]{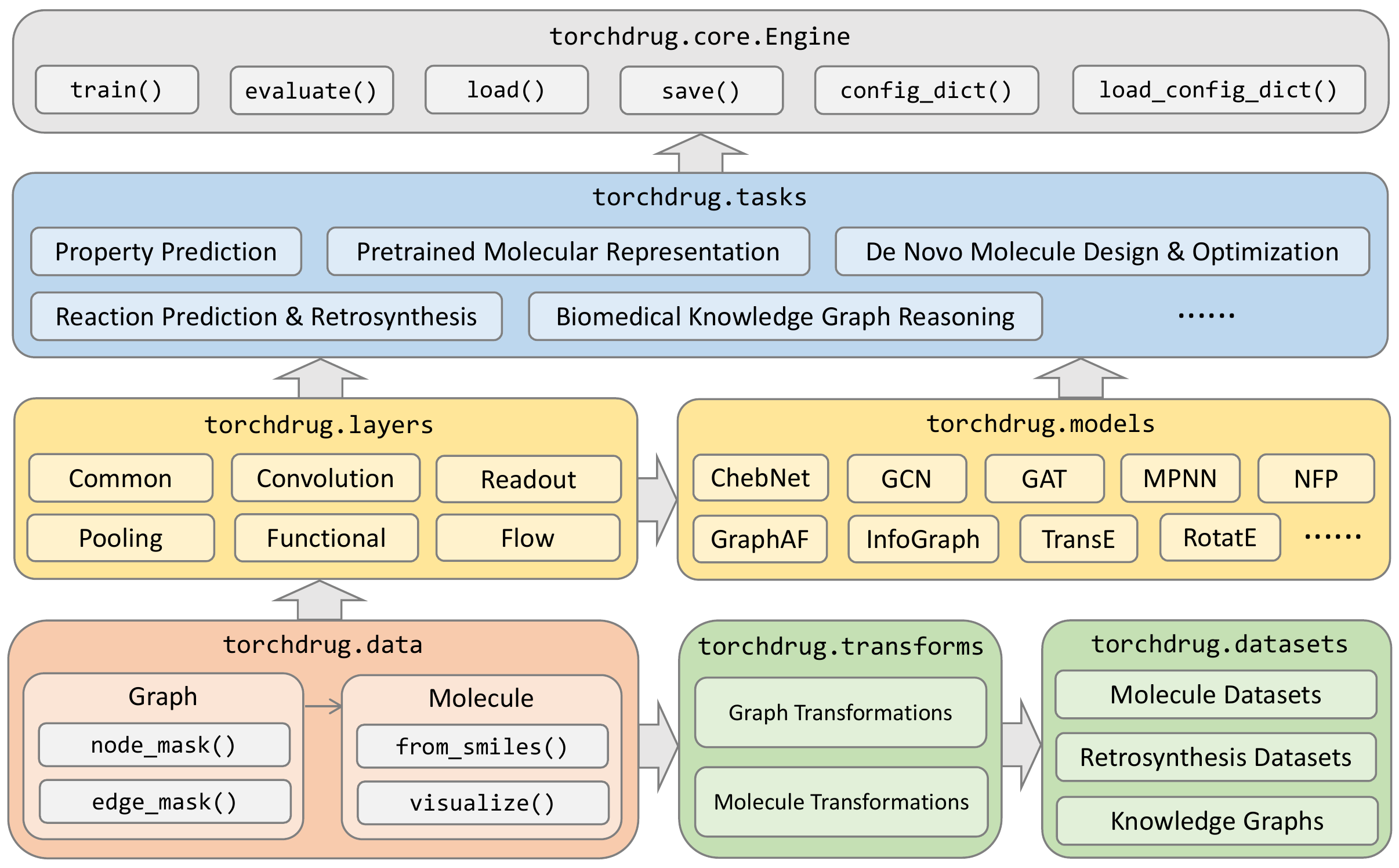}
    \vspace{-1.2em}
    \caption{Overview of the hierarchy of \library.}
    \vspace{-1em}
    \label{fig:overview}
\end{figure}

To accelerate the process of drug discovery through machine learning, we see a critical need to develop an open-source machine learning platform for drug discovery. Here we present such a platform, called \library. \library provides a hierarchical interface to accommodate different demands in the development of drug discovery. At the low level, \library encapsulates graphs and molecules as basic data structures, and provides GPU-accelerated graph operations, along with standard datasets in a PyTorch-style interface. At the mid-level, \library supplies popular building blocks of graph representation learning models (e.g., MPNN~\citep{gilmer2017neural}), which can be used to quickly construct models for drug discovery. The high level contains reusable routines for a variety of important tasks in drug discovery, ranging from molecular property prediction, pretrained molecular representations, de novo molecule design and optimization, retrosynthesis prediction to biomedical knowledge graph reasoning (e.g., for drug repurposing). Figure~\ref{fig:overview} presents an overview of the \library library. \library has received more than 5,000 downloads on PyPI and Anaconda since its first release in August 2021.
\section{Existing Systems and Implementations}

\noindent \textbf{Graph Machine Learning Systems.}
The most prominent packages are PyTorch-Geometric (PyG)~\citep{fey2019fast} and Deep Graph Library (DGL)~\citep{wang2019deep}, which are targeted at building graph neural networks (GNNs) in PyTorch~\citep{paszke2019pytorch}. DGL additionally supports MXNet~\citep{chen2015mxnet} backend. Other similar packages include GraphNets~\citep{battaglia2018relational}, StellarGraph~\citep{StellarGraph}, Spektral~\citep{grattarola2020graph}, tf\_geometric~\citep{hu2021efficient} for Tensorflow~\citep{abadi2016tensorflow}, CogDL~\citep{cen2021cogdl} for PyTorch, and Jraph~\citep{jraph2020github} for JAX~\citep{jax2018github}. All these libraries cover the low-level graph operations (e.g., batch of graphs) and mid-level models (e.g., GIN~\citep{xu2018powerful}) for GNN architectures. By contrast, DIG~\citep{liu2021dig} focuses on high-level tasks on graphs, such as graph generation and self-supervised learning. To our best knowledge, there is no general graph machine learning system targeting at the research and development need of drug discovery tasks.

\noindent \textbf{Drug Discovery Implementations.}
Another stream of software is dedicated to the implementation of drug discovery tasks.
For example, DeepChem~\citep{ramsundar2019deep} provides standard datasets and models for molecule property prediction. \citet{you2018graph} implements a codebase for conducting research on de novo molecular design. \citet{dai2019retrosynthesis, shi*2021confgf} develop two delicate packages for retrosynthesis prediction and 3D molecular conformation prediction, respectively. However, these efforts are mostly targeted on a specific task with a limited range of models, which restricts the development and benchmarking of new fundamental models. Other implementations like Therapeutics Data Commons (TDC)~\citep{huang2021therapeutics} and ATOM3D~\citep{townshend2020atom3d} focus on datasets and evaluation toolkits for drug discovery tasks, but lack implementation of models.

By comparison, \library is a graph machine learning system for drug discovery, with flexible interface for low-level operations, mid-level models and high-level tasks.
\section{Key Features}

\library offers two key features: 1) low-level data structures and graph operations that can be manipulated with minimal domain knowledge and GPU acceleration. 2) mid-level datasets, layers, models and high-level tasks that support rapid prototyping of ideas.

\subsection{Data Structures and Graph Operations}

The core data structures of \library are homogeneous graphs, knowledge graphs (together in \code{data.Graph}), and molecules (\code{data.Molecule}). Like tensors in PyTorch, these data structures are designed to be the first-class citizen in \library, and serve as input and output of many functions. Our data structures support a lot of graph operations, such as node masking (\code{data.Graph.node\_mask}), extracting connected components (\code{data.Graph.connected\_components}) and converting ions to molecules (\code{data.Molecule.ion\_to\_molecule}), as well as their batched variants. All the graph operations are implemented based on standard PyTorch operations, which support auto differentation and can be seamlessly switched between CPUs and GPUs. For example, the following code snippet creates a batch of 4 molecules, sends it to a GPU, repeats the batch and visualizes the results.
\begin{figure}[!h]
    \centering
    \vspace{-1em}
    \input{codes/graph_operation}
    \includegraphics[width=0.95\textwidth]{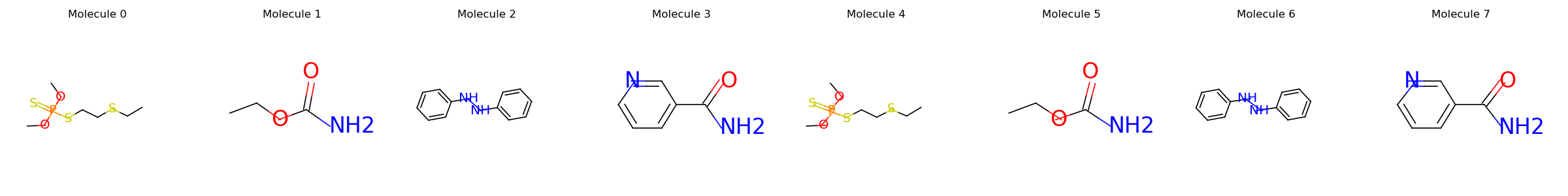}
    \vspace{-1.5em}
\end{figure}

The data structures also contain several predefined node-level, edge-level and graph-level attributes that are useful for building machine learning models. For example, the type of atoms in a molecule may be used as an input feature to some property prediction model. Users may also register arbitrary attributes depending on their tasks. All the attributes are automatically maintained in all of our graph operations.

\subsection{Datasets, Layers, Models and Tasks}

\noindent \textbf{Datasets.}
The \code{datasets} module provides 30 common datasets for 5 drug discovery tasks.
These datasets inherit the \code{Dataset} class from PyTorch and further provide data loading and \code{\_\_getitem\_\_} functions, which facilitates the interaction with dataloaders in PyTorch.

\noindent \textbf{Layers and Models.}
The \code{layers} and \code{models} modules implement layers and models for representation learning respectively. This lets users switch between standard models or custom models from standard layers.
Our interface follows the convention in PyTorch, which minimizes the cognitive load of users.
Classes in layers (e.g., \code{GCNConv}) are similar to the layers in \code{torch.nn}, while classes in \code{models} (e.g., \code{GCN}) are similar to \code{torchvision.models}.

\noindent \textbf{Tasks.}
The \code{tasks} module contains high-level routines of machine learning tasks in drug discovery. 
Typically, these include dataset preprocessing, prediction, training and evaluation. 
Each task is abstracted as a model-agnostic class in tasks, which can be used with any basic representation learning models (e.g., GIN~\citep{xu2018powerful}). 
Currently, \library supports 5 tasks: property prediction, pretrained molecular representations, de novo molecule design, retrosynthesis and biomedical knowledge graph reasoning. 
A full list of tasks and models supported by \library are showed in Table~\ref{tab:task}.

\begin{table}[!h]
    \centering
    \vspace{-1em}
    \caption{Drug discovery tasks and models supported by \library.}
    \begin{adjustbox}{max width=\textwidth}
        \begin{tabular}{lllll}
            \toprule
            \bf{Task} & \bf{Model} & \\
            \midrule
            \multirow{3}{*}{\bf{Property Prediction}}
            & Neural Fingerprint~\citep{duvenaud2015convolutional} & ChebyNet~\citep{defferrard2016convolutional} & GCN~\citep{kipf2016semi} \\
            & ENN-S2S~\citep{gilmer2017neural} & SchNet~\citep{schutt2017schnet} & GAT~\citep{velivckovic2017graph} \\
            & RGCN~\citep{schlichtkrull2018modeling} & GIN~\citep{xu2018powerful} \\
            \midrule
            \multirow{2}{*}{\bf{\shortstack[l]{Pretrained Molecular\\Representation}}}
            & InfoGraph~\citep{sun2019infograph} & Edge Prediction~\citep{hamilton2017inductive} & Attribute Masking~\citep{hu2019strategies} \\
            & Context Prediction~\citep{hu2019strategies} \\
            \midrule
            \bf{De Novo Molecule Design and Optimization}
            & GCPN~\citep{you2018graph} & GraphAF~\citep{shi2020graphaf} \\
            \midrule
            \bf{Retrosynthesis Prediction}
            & G2Gs~\citep{shi2020graph} \\
            \midrule
            \multirow{3}{*}{\bf{\shortstack[l]{Biomedical\\Knowledge Graph\\Reasoning}}}
            & TransE~\citep{bordes2013translating} & DistMult~\citep{yang2014embedding} & ComplEx~\citep{trouillon2016complex} \\ 
            & NeuralLP~\citep{yang2017differentiable} & SimplE~\citep{kazemi2018simple} & RotatE~\citep{sun2019rotate} \\
            & KBGAT~\citep{nathani2019learning} \\
            \bottomrule
        \end{tabular}
    \end{adjustbox}
    \vspace{-1.5em}
    \label{tab:task}
\end{table}
\section{Conclusion}

We present \library, a powerful and flexible machine learning platform for drug discovery. It has the potential to accelerate the research and development of drug discovery. In the future, we plan to further add tasks and models for protein representation learning.

\vskip 0.2in
\bibliography{reference}

\begin{thebibliography}{46}
\providecommand{\natexlab}[1]{#1}
\providecommand{\url}[1]{\texttt{#1}}
\expandafter\ifx\csname urlstyle\endcsname\relax
  \providecommand{\doi}[1]{doi: #1}\else
  \providecommand{\doi}{doi: \begingroup \urlstyle{rm}\Url}\fi

\bibitem[Abadi et~al.(2016)Abadi, Barham, Chen, Chen, Davis, Dean, Devin,
  Ghemawat, Irving, Isard, et~al.]{abadi2016tensorflow}
Mart{\'\i}n Abadi, Paul Barham, Jianmin Chen, Zhifeng Chen, Andy Davis, Jeffrey
  Dean, Matthieu Devin, Sanjay Ghemawat, Geoffrey Irving, Michael Isard, et~al.
\newblock Tensorflow: A system for large-scale machine learning.
\newblock In \emph{12th $\{$USENIX$\}$ symposium on operating systems design
  and implementation ($\{$OSDI$\}$ 16)}, pages 265--283, 2016.

\bibitem[Baek et~al.(2021)Baek, DiMaio, Anishchenko, Dauparas, Ovchinnikov,
  Lee, Wang, Cong, Kinch, Schaeffer, et~al.]{baek2021accurate}
Minkyung Baek, Frank DiMaio, Ivan Anishchenko, Justas Dauparas, Sergey
  Ovchinnikov, Gyu~Rie Lee, Jue Wang, Qian Cong, Lisa~N Kinch, R~Dustin
  Schaeffer, et~al.
\newblock Accurate prediction of protein structures and interactions using a
  three-track neural network.
\newblock \emph{Science}, 373\penalty0 (6557):\penalty0 871--876, 2021.

\bibitem[Battaglia et~al.(2018)Battaglia, Hamrick, Bapst, Sanchez-Gonzalez,
  Zambaldi, Malinowski, Tacchetti, Raposo, Santoro, Faulkner,
  et~al.]{battaglia2018relational}
Peter~W Battaglia, Jessica~B Hamrick, Victor Bapst, Alvaro Sanchez-Gonzalez,
  Vinicius Zambaldi, Mateusz Malinowski, Andrea Tacchetti, David Raposo, Adam
  Santoro, Ryan Faulkner, et~al.
\newblock Relational inductive biases, deep learning, and graph networks.
\newblock \emph{arXiv preprint arXiv:1806.01261}, 2018.

\bibitem[Bordes et~al.(2013)Bordes, Usunier, Garcia-Duran, Weston, and
  Yakhnenko]{bordes2013translating}
Antoine Bordes, Nicolas Usunier, Alberto Garcia-Duran, Jason Weston, and Oksana
  Yakhnenko.
\newblock Translating embeddings for modeling multi-relational data.
\newblock In \emph{Neural Information Processing Systems (NIPS)}, pages 1--9,
  2013.

\bibitem[Bradbury et~al.(2018)Bradbury, Frostig, Hawkins, Johnson, Leary,
  Maclaurin, Necula, Paszke, Vander{P}las, Wanderman-{M}ilne, and
  Zhang]{jax2018github}
James Bradbury, Roy Frostig, Peter Hawkins, Matthew~James Johnson, Chris Leary,
  Dougal Maclaurin, George Necula, Adam Paszke, Jake Vander{P}las, Skye
  Wanderman-{M}ilne, and Qiao Zhang.
\newblock {JAX}: composable transformations of {P}ython+{N}um{P}y programs,
  2018.
\newblock URL \url{http://github.com/google/jax}.

\bibitem[Bradshaw et~al.(2018)Bradshaw, Kusner, Paige, Segler, and
  Hern{\'a}ndez-Lobato]{bradshaw2018generative}
John Bradshaw, Matt~J Kusner, Brooks Paige, Marwin~HS Segler, and
  Jos{\'e}~Miguel Hern{\'a}ndez-Lobato.
\newblock A generative model for electron paths.
\newblock \emph{arXiv preprint arXiv:1805.10970}, 2018.

\bibitem[Cen et~al.(2021)Cen, Hou, Wang, Chen, Luo, Yao, Zeng, Guo, Zhang, Dai,
  et~al.]{cen2021cogdl}
Yukuo Cen, Zhenyu Hou, Yan Wang, Qibin Chen, Yizhen Luo, Xingcheng Yao, Aohan
  Zeng, Shiguang Guo, Peng Zhang, Guohao Dai, et~al.
\newblock Cogdl: An extensive toolkit for deep learning on graphs.
\newblock \emph{arXiv preprint arXiv:2103.00959}, 2021.

\bibitem[Chen et~al.(2015)Chen, Li, Li, Lin, Wang, Wang, Xiao, Xu, Zhang, and
  Zhang]{chen2015mxnet}
Tianqi Chen, Mu~Li, Yutian Li, Min Lin, Naiyan Wang, Minjie Wang, Tianjun Xiao,
  Bing Xu, Chiyuan Zhang, and Zheng Zhang.
\newblock Mxnet: A flexible and efficient machine learning library for
  heterogeneous distributed systems.
\newblock \emph{arXiv preprint arXiv:1512.01274}, 2015.

\bibitem[Dai et~al.(2019)Dai, Li, Coley, Dai, and Song]{dai2019retrosynthesis}
Hanjun Dai, Chengtao Li, Connor Coley, Bo~Dai, and Le~Song.
\newblock Retrosynthesis prediction with conditional graph logic network.
\newblock In \emph{Advances in Neural Information Processing Systems}, pages
  8870--8880, 2019.

\bibitem[Data61(2018)]{StellarGraph}
CSIRO's Data61.
\newblock Stellargraph machine learning library.
\newblock \url{https://github.com/stellargraph/stellargraph}, 2018.

\bibitem[Defferrard et~al.(2016)Defferrard, Bresson, and
  Vandergheynst]{defferrard2016convolutional}
Micha{\"e}l Defferrard, Xavier Bresson, and Pierre Vandergheynst.
\newblock Convolutional neural networks on graphs with fast localized spectral
  filtering.
\newblock \emph{Advances in neural information processing systems},
  29:\penalty0 3844--3852, 2016.

\bibitem[DiMasi et~al.(2016)DiMasi, Grabowski, and
  Hansen]{dimasi2016innovation}
Joseph~A DiMasi, Henry~G Grabowski, and Ronald~W Hansen.
\newblock Innovation in the pharmaceutical industry: new estimates of r\&d
  costs.
\newblock \emph{Journal of health economics}, 47:\penalty0 20--33, 2016.

\bibitem[Duvenaud et~al.(2015)Duvenaud, Maclaurin, Iparraguirre, Bombarell,
  Hirzel, Aspuru-Guzik, and Adams]{duvenaud2015convolutional}
David~K Duvenaud, Dougal Maclaurin, Jorge Iparraguirre, Rafael Bombarell,
  Timothy Hirzel, Alan Aspuru-Guzik, and Ryan~P Adams.
\newblock Convolutional networks on graphs for learning molecular fingerprints.
\newblock \emph{Advances in Neural Information Processing Systems},
  28:\penalty0 2224--2232, 2015.

\bibitem[Fey and Lenssen(2019)]{fey2019fast}
Matthias Fey and Jan~Eric Lenssen.
\newblock Fast graph representation learning with pytorch geometric.
\newblock \emph{arXiv preprint arXiv:1903.02428}, 2019.

\bibitem[Gilmer et~al.(2017)Gilmer, Schoenholz, Riley, Vinyals, and
  Dahl]{gilmer2017neural}
Justin Gilmer, Samuel~S Schoenholz, Patrick~F Riley, Oriol Vinyals, and
  George~E Dahl.
\newblock Neural message passing for quantum chemistry.
\newblock In \emph{International Conference on Machine Learning}, pages
  1263--1272. PMLR, 2017.

\bibitem[Godwin* et~al.(2020)Godwin*, Keck*, Battaglia, Bapst, Kipf, Li,
  Stachenfeld, Veli\v{c}kovi\'{c}, and Sanchez-Gonzalez]{jraph2020github}
Jonathan Godwin*, Thomas Keck*, Peter Battaglia, Victor Bapst, Thomas Kipf,
  Yujia Li, Kimberly Stachenfeld, Petar Veli\v{c}kovi\'{c}, and Alvaro
  Sanchez-Gonzalez.
\newblock {J}raph: {A} library for graph neural networks in jax., 2020.
\newblock URL \url{http://github.com/deepmind/jraph}.

\bibitem[Grattarola and Alippi(2020)]{grattarola2020graph}
Daniele Grattarola and Cesare Alippi.
\newblock Graph neural networks in tensorflow and keras with spektral.
\newblock \emph{arXiv preprint arXiv:2006.12138}, 2020.

\bibitem[Hamilton et~al.(2017)Hamilton, Ying, and
  Leskovec]{hamilton2017inductive}
William~L Hamilton, Rex Ying, and Jure Leskovec.
\newblock Inductive representation learning on large graphs.
\newblock \emph{arXiv preprint arXiv:1706.02216}, 2017.

\bibitem[Hu et~al.(2021)Hu, Qian, Fang, Wang, Zhao, Zhang, and
  Xu]{hu2021efficient}
Jun Hu, Shengsheng Qian, Quan Fang, Youze Wang, Quan Zhao, Huaiwen Zhang, and
  Changsheng Xu.
\newblock Efficient graph deep learning in tensorflow with tf\_geometric.
\newblock \emph{arXiv preprint arXiv:2101.11552}, 2021.

\bibitem[Hu et~al.(2019)Hu, Liu, Gomes, Zitnik, Liang, Pande, and
  Leskovec]{hu2019strategies}
Weihua Hu, Bowen Liu, Joseph Gomes, Marinka Zitnik, Percy Liang, Vijay Pande,
  and Jure Leskovec.
\newblock Strategies for pre-training graph neural networks.
\newblock In \emph{International Conference on Learning Representations}, 2019.

\bibitem[Huang et~al.(2021)Huang, Fu, Gao, Zhao, Roohani, Leskovec, Coley,
  Xiao, Sun, and Zitnik]{huang2021therapeutics}
Kexin Huang, Tianfan Fu, Wenhao Gao, Yue Zhao, Yusuf Roohani, Jure Leskovec,
  Connor~W Coley, Cao Xiao, Jimeng Sun, and Marinka Zitnik.
\newblock Therapeutics data commons: machine learning datasets and tasks for
  therapeutics.
\newblock \emph{arXiv preprint arXiv:2102.09548}, 2021.

\bibitem[Jin et~al.(2017)Jin, Coley, Barzilay, and Jaakkola]{jin2017predicting}
Wengong Jin, Connor~W Coley, Regina Barzilay, and Tommi Jaakkola.
\newblock Predicting organic reaction outcomes with weisfeiler-lehman network.
\newblock \emph{arXiv preprint arXiv:1709.04555}, 2017.

\bibitem[Jumper et~al.(2021)Jumper, Evans, Pritzel, Green, Figurnov,
  Ronneberger, Tunyasuvunakool, Bates, {\v{Z}}{\'\i}dek, Potapenko,
  et~al.]{jumper2021highly}
John Jumper, Richard Evans, Alexander Pritzel, Tim Green, Michael Figurnov,
  Olaf Ronneberger, Kathryn Tunyasuvunakool, Russ Bates, Augustin
  {\v{Z}}{\'\i}dek, Anna Potapenko, et~al.
\newblock Highly accurate protein structure prediction with alphafold.
\newblock \emph{Nature}, 596\penalty0 (7873):\penalty0 583--589, 2021.

\bibitem[Kazemi and Poole(2018)]{kazemi2018simple}
Seyed~Mehran Kazemi and David Poole.
\newblock Simple embedding for link prediction in knowledge graphs.
\newblock \emph{arXiv preprint arXiv:1802.04868}, 2018.

\bibitem[Kipf and Welling(2016)]{kipf2016semi}
Thomas~N Kipf and Max Welling.
\newblock Semi-supervised classification with graph convolutional networks.
\newblock \emph{arXiv preprint arXiv:1609.02907}, 2016.

\bibitem[Liu et~al.(2021)Liu, Luo, Wang, Xie, Yuan, Gui, Yu, Xu, Zhang, Liu,
  Yan, Liu, Fu, Oztekin, Zhang, and Ji]{liu2021dig}
Meng Liu, Youzhi Luo, Limei Wang, Yaochen Xie, Hao Yuan, Shurui Gui, Haiyang
  Yu, Zhao Xu, Jingtun Zhang, Yi~Liu, Keqiang Yan, Haoran Liu, Cong Fu, Bora~M
  Oztekin, Xuan Zhang, and Shuiwang Ji.
\newblock Dig: A turnkey library for diving into graph deep learning research.
\newblock \emph{Journal of Machine Learning Research}, 22\penalty0
  (240):\penalty0 1--9, 2021.
\newblock URL \url{http://jmlr.org/papers/v22/21-0343.html}.

\bibitem[Nathani et~al.(2019)Nathani, Chauhan, Sharma, and
  Kaul]{nathani2019learning}
Deepak Nathani, Jatin Chauhan, Charu Sharma, and Manohar Kaul.
\newblock Learning attention-based embeddings for relation prediction in
  knowledge graphs.
\newblock \emph{arXiv preprint arXiv:1906.01195}, 2019.

\bibitem[Paszke et~al.(2019)Paszke, Gross, Massa, Lerer, Bradbury, Chanan,
  Killeen, Lin, Gimelshein, Antiga, et~al.]{paszke2019pytorch}
Adam Paszke, Sam Gross, Francisco Massa, Adam Lerer, James Bradbury, Gregory
  Chanan, Trevor Killeen, Zeming Lin, Natalia Gimelshein, Luca Antiga, et~al.
\newblock Pytorch: An imperative style, high-performance deep learning library.
\newblock \emph{Advances in neural information processing systems},
  32:\penalty0 8026--8037, 2019.

\bibitem[Ramsundar et~al.(2019)Ramsundar, Eastman, Walters, Pande, Leswing, and
  Wu]{ramsundar2019deep}
Bharath Ramsundar, Peter Eastman, Patrick Walters, Vijay Pande, Karl Leswing,
  and Zhenqin Wu.
\newblock \emph{Deep Learning for the Life Sciences}.
\newblock O'Reilly Media, 2019.
\newblock
  \url{https://www.amazon.com/Deep-Learning-Life-Sciences-Microscopy/dp/1492039837}.

\bibitem[Schlichtkrull et~al.(2018)Schlichtkrull, Kipf, Bloem, Van Den~Berg,
  Titov, and Welling]{schlichtkrull2018modeling}
Michael Schlichtkrull, Thomas~N Kipf, Peter Bloem, Rianne Van Den~Berg, Ivan
  Titov, and Max Welling.
\newblock Modeling relational data with graph convolutional networks.
\newblock In \emph{European semantic web conference}, pages 593--607. Springer,
  2018.

\bibitem[Sch{\"u}tt et~al.(2017)Sch{\"u}tt, Kindermans, Sauceda, Chmiela,
  Tkatchenko, and M{\"u}ller]{schutt2017schnet}
Kristof~T Sch{\"u}tt, Pieter-Jan Kindermans, Huziel~E Sauceda, Stefan Chmiela,
  Alexandre Tkatchenko, and Klaus-Robert M{\"u}ller.
\newblock Schnet: A continuous-filter convolutional neural network for modeling
  quantum interactions.
\newblock \emph{arXiv preprint arXiv:1706.08566}, 2017.

\bibitem[Shi et~al.(2020{\natexlab{a}})Shi, Xu, Guo, Zhang, and
  Tang]{shi2020graph}
Chence Shi, Minkai Xu, Hongyu Guo, Ming Zhang, and Jian Tang.
\newblock A graph to graphs framework for retrosynthesis prediction.
\newblock In \emph{International Conference on Machine Learning}, pages
  8818--8827. PMLR, 2020{\natexlab{a}}.

\bibitem[Shi et~al.(2020{\natexlab{b}})Shi, Xu, Zhu, Zhang, Zhang, and
  Tang]{shi2020graphaf}
Chence Shi, Minkai Xu, Zhaocheng Zhu, Weinan Zhang, Ming Zhang, and Jian Tang.
\newblock Graphaf: a flow-based autoregressive model for molecular graph
  generation.
\newblock \emph{International Conference on Learning Representations},
  2020{\natexlab{b}}.

\bibitem[Shi et~al.(2021)Shi, Luo, Xu, and Tang]{shi*2021confgf}
Chence Shi, Shitong Luo, Minkai Xu, and Jian Tang.
\newblock Learning gradient fields for molecular conformation generation.
\newblock In \emph{International Conference on Machine Learning}, 2021.

\bibitem[Sun et~al.(2019{\natexlab{a}})Sun, Hoffmann, Verma, and
  Tang]{sun2019infograph}
Fan-Yun Sun, Jordan Hoffmann, Vikas Verma, and Jian Tang.
\newblock Infograph: Unsupervised and semi-supervised graph-level
  representation learning via mutual information maximization.
\newblock \emph{arXiv preprint arXiv:1908.01000}, 2019{\natexlab{a}}.

\bibitem[Sun et~al.(2019{\natexlab{b}})Sun, Deng, Nie, and Tang]{sun2019rotate}
Zhiqing Sun, Zhi-Hong Deng, Jian-Yun Nie, and Jian Tang.
\newblock Rotate: Knowledge graph embedding by relational rotation in complex
  space.
\newblock \emph{arXiv preprint arXiv:1902.10197}, 2019{\natexlab{b}}.

\bibitem[Townshend et~al.(2020)Townshend, V{\"o}gele, Suriana, Derry, Powers,
  Laloudakis, Balachandar, Jing, Anderson, Eismann,
  et~al.]{townshend2020atom3d}
Raphael~JL Townshend, Martin V{\"o}gele, Patricia Suriana, Alexander Derry,
  Alexander Powers, Yianni Laloudakis, Sidhika Balachandar, Bowen Jing, Brandon
  Anderson, Stephan Eismann, et~al.
\newblock Atom3d: Tasks on molecules in three dimensions.
\newblock \emph{arXiv preprint arXiv:2012.04035}, 2020.

\bibitem[Trouillon et~al.(2016)Trouillon, Welbl, Riedel, Gaussier, and
  Bouchard]{trouillon2016complex}
Th{\'e}o Trouillon, Johannes Welbl, Sebastian Riedel, {\'E}ric Gaussier, and
  Guillaume Bouchard.
\newblock Complex embeddings for simple link prediction.
\newblock In \emph{International Conference on Machine Learning}, pages
  2071--2080. PMLR, 2016.

\bibitem[Veli{\v{c}}kovi{\'c} et~al.(2017)Veli{\v{c}}kovi{\'c}, Cucurull,
  Casanova, Romero, Lio, and Bengio]{velivckovic2017graph}
Petar Veli{\v{c}}kovi{\'c}, Guillem Cucurull, Arantxa Casanova, Adriana Romero,
  Pietro Lio, and Yoshua Bengio.
\newblock Graph attention networks.
\newblock \emph{arXiv preprint arXiv:1710.10903}, 2017.

\bibitem[Wang et~al.(2019)Wang, Yu, Zheng, Gan, Gai, Ye, Li, Zhou, Huang, Ma,
  et~al.]{wang2019deep}
Minjie Wang, Lingfan Yu, Da~Zheng, Quan Gan, Yu~Gai, Zihao Ye, Mufei Li,
  Jinjing Zhou, Qi~Huang, Chao Ma, et~al.
\newblock Deep graph library: Towards efficient and scalable deep learning on
  graphs.
\newblock 2019.

\bibitem[Wang et~al.(2020)Wang, Li, Wang, Parulian, Han, Ma, Tu, Lin, Zhang,
  Liu, et~al.]{wang2020covid}
Qingyun Wang, Manling Li, Xuan Wang, Nikolaus Parulian, Guangxing Han, Jiawei
  Ma, Jingxuan Tu, Ying Lin, Haoran Zhang, Weili Liu, et~al.
\newblock Covid-19 literature knowledge graph construction and drug repurposing
  report generation.
\newblock \emph{arXiv preprint arXiv:2007.00576}, 2020.

\bibitem[Xu et~al.(2018)Xu, Hu, Leskovec, and Jegelka]{xu2018powerful}
Keyulu Xu, Weihua Hu, Jure Leskovec, and Stefanie Jegelka.
\newblock How powerful are graph neural networks?
\newblock \emph{arXiv preprint arXiv:1810.00826}, 2018.

\bibitem[Yang et~al.(2014)Yang, Yih, He, Gao, and Deng]{yang2014embedding}
Bishan Yang, Wen-tau Yih, Xiaodong He, Jianfeng Gao, and Li~Deng.
\newblock Embedding entities and relations for learning and inference in
  knowledge bases.
\newblock \emph{arXiv preprint arXiv:1412.6575}, 2014.

\bibitem[Yang et~al.(2017)Yang, Yang, and Cohen]{yang2017differentiable}
Fan Yang, Zhilin Yang, and William~W Cohen.
\newblock Differentiable learning of logical rules for knowledge base
  reasoning.
\newblock \emph{arXiv preprint arXiv:1702.08367}, 2017.

\bibitem[You et~al.(2018)You, Liu, Ying, Pande, and Leskovec]{you2018graph}
Jiaxuan You, Bowen Liu, Zhitao Ying, Vijay Pande, and Jure Leskovec.
\newblock Graph convolutional policy network for goal-directed molecular graph
  generation.
\newblock \emph{Advances in Neural Information Processing Systems}, 31, 2018.

\bibitem[Zhao et~al.(2020)Zhao, Qin, Liu, and Wang]{zhao2020biomedical}
Sendong Zhao, Bing Qin, Ting Liu, and Fei Wang.
\newblock Biomedical knowledge graph refinement with embedding and logic rules.
\newblock \emph{arXiv preprint arXiv:2012.01031}, 2020.

\end{thebibliography}

\end{document}